\documentclass[10pt,leqno]{amsart}
\usepackage[margin=0.78in]{geometry}
\usepackage[english]{babel}
\usepackage{times}
\usepackage{microtype}
\usepackage{booktabs}
\usepackage{graphicx}
\usepackage{amsmath}
\usepackage{amssymb}
\usepackage{hyperref}
\usepackage[numbers,sort&compress]{natbib}

\hypersetup{colorlinks=true,linkcolor=blue,citecolor=blue,urlcolor=blue}

\title{\vspace{-1.2em}RASC+: Retrieval-Constrained LLM Adjudication for Clinical Value Set Authoring}
\author{Sumit Mukherjee\\Oracle Health, USA}
\date{}

\begin{document}
\maketitle

\begin{abstract}
Clinical value sets define the standardized terminology codes used in quality measurement, phenotyping, cohort construction, and clinical decision support. The recently introduced Retrieval-Augmented Set Completion (RASC) benchmark showed that direct zero-shot large language model (LLM) generation is poorly suited to this task: clinical code systems are large, version-controlled, and not reliably memorized by language models. We study a stage-wise alternative in which candidate-pool construction is optimized for recall and a constrained LLM adjudicator is optimized for candidate selection. On the full 3,744-value-set RASC test split, Qwen3-based retrieval with vocabulary-aware expansion and code-display rescue retrieval increases candidate-pool recall from the original RASC retrieval baseline of 0.553 to 0.730; on the held-out-publisher stratum, pool recall is 0.655. The higher-recall pool alone is not sufficient: applying the original SAPBert cross-encoder to this expanded pool gives full-test macro F1 of 0.287 and held-out-publisher macro F1 of 0.233. Replacing the stage-2 selector with blinded GPT-5 adjudication over the same pool increases full-test macro F1 to 0.549 and held-out-publisher macro F1 to 0.533. These results show that retrieval-constrained LLM adjudication can substantially improve value set completion while preserving the safety constraint that all returned codes must come from an auditable candidate pool.
\end{abstract}

\section{Introduction}
Clinical value sets specify which terminology codes operationalize a clinical concept. They are used throughout electronic quality measurement, phenotyping, cohort definition, and decision support. Authoring a value set is not ordinary text generation: the output is a subset of a large, versioned terminology universe spanning systems such as SNOMED-CT, ICD-10-CM, RxNorm, LOINC, CPT, and others. A useful system must be both complete enough to recover relevant codes and precise enough to avoid overwhelming human reviewers with plausible but incorrect candidates.

Mukherjee et al.~\citep{mukherjee2026retrieveclassifycorpusgroundedautomation} introduced Retrieval-Augmented Set Completion (RASC), a method for clinical value set authoring built from publicly available Value Set Authority Center (VSAC) value sets~\citep{nlmvsac}. RASC's central design is to retrieve similar prior value sets, union their codes into a candidate pool, and then classify each candidate. This framing avoids unconstrained code generation and makes the retrieval stage an explicit upper bound on downstream recall. The original RASC paper reported that a SAPBert cross-encoder achieved value-set-level macro F1 0.298, outperforming a zero-shot GPT-4o generator that achieved macro F1 0.105 and frequently returned codes absent from VSAC.

This work asks whether the same retrieve-then-select formulation can be strengthened by optimizing each stage for its own objective. We treat candidate-pool construction and candidate adjudication as separate problems. The retrieval stage should maximize recall subject to a reviewable pool size, even if that lowers pool precision. The adjudication stage should then recover precision while preserving as much of the pool recall as possible. This decomposition is different from direct LLM generation, which asks a model to recall exact clinical codes from a large universe. Retrieval-constrained adjudication instead asks whether each supplied code belongs in a target value set, preserving auditability and corpus grounding while using an LLM only as a bounded selector.

We make two empirical contributions. First, we show that a combination of Qwen3-based semantic retrieval, vocabulary-aware graph expansion, and code-display rescue retrieval raises full-test candidate-pool recall from the RASC baseline of 0.553 to 0.730, including 0.655 on held-out publishers — establishing a substantially higher recall ceiling for any downstream selector. Second, we show that GPT-5 adjudication under a leakage-controlled protocol converts this larger, noisier pool into end-to-end improvements that the original SAPBert cross-encoder cannot: full-test macro F1 rises from 0.287 to 0.549, with OOD (value set publishers not seen in training) macro F1 rising from 0.233 to 0.533 on the same pool. 

\section{Related Work}
\subsection{Retrieval-Augmented Set Completion}
RASC formalizes value set authoring as set completion over a large discrete terminology universe~\citep{mukherjee2026retrieveclassifycorpusgroundedautomation}. The benchmark contains 11,803 VSAC value sets after filtering and uses a held-out publisher split: Clinical Architecture and CSTE Steward are withheld entirely for test. This split is important because it measures generalization to authoring styles and concept distributions not seen during training. RASC also reports both retrieval recall and downstream classifier metrics, allowing failures to be attributed either to absent candidates or to incorrect candidate selection. We use that decomposition directly, while studying a different pipeline design: a higher-recall pool constructor followed by constrained LLM adjudication. The original RASC manuscript reports the primary full-test baselines used here; it does not provide the same ID/OOD breakdown for the original SAPBert cross-encoder, so OOD stage-2 comparisons in this paper are made against SAPBert applied to the expanded pool.

\subsection{Retrieval-Constrained Generation and Adjudication}
Retrieval-augmented generation typically supplies documents as context for an unconstrained generator~\citep{lewis2020retrieval}. RASC differs because retrieval supplies the feasible output set itself. The downstream model is not merely grounded by retrieved evidence; it is constrained to return a subset of retrieved items. This distinction is central for clinical value set authoring, where an invalid code is not a harmless paraphrase but an unusable artifact. Our GPT-5 experiment follows the same constraint: the LLM selects candidate identifiers from a supplied pool and is never asked to generate codes de novo.

\subsection{Biomedical Retrieval Models}
The original RASC retrieval stage used SAPBert title embeddings. SAPBert is a biomedical entity representation model trained with UMLS synonym supervision~\citep{liu2021sapbert}, and related biomedical retrieval systems such as MedCPT demonstrate the value of domain-specific contrastive pretraining for biomedical search~\citep{jin2023medcpt}. Our pool-construction experiments use the Qwen3-Embedding-0.6B checkpoint, a member of the Qwen3 embedding family designed for text embedding and reranking~\citep{zhang2025qwen3embedding}. Dense retrieval uses normalized embeddings indexed with FAISS~\citep{douze2024faiss}. The final LLM evaluation uses the highest-recall pool configuration available in these experiments: Qwen3 retrieval, vocabulary-aware expansion, and code-display rescue retrieval.

\section{Methods}
\subsection{VSAC Corpus and Test Split}
We use the VSAC-derived benchmark and split design introduced by RASC~\citep{mukherjee2026retrieveclassifycorpusgroundedautomation}. Each example is a value set expansion consisting of a title, metadata, terminology systems, and a gold set of \((\mathrm{code}, \mathrm{system})\) pairs. We do not re-create the corpus or alter the split. The full test split contains 3,744 value sets: 1,423 from non-held-out publishers and 2,321 from held-out publishers. The held-out publishers are Clinical Architecture and CSTE Steward, matching the original RASC split. All candidate-pool and adjudication results are reported on this full test split unless explicitly stated otherwise.

\subsection{Semantic Retrieval}
Let \(\mathcal{V}_{\mathrm{train}}\) denote the corpus of source value sets available for retrieval, and let \(U\) denote the universe of observed \((\mathrm{code}, \mathrm{system})\) pairs. For a target value set \(v\) with title \(t_v\), the goal of stage 1 is to construct a candidate pool \(C_v \subset U\) with high coverage of the unknown target set \(Y_v\). This objective differs from final prediction: candidate-pool construction is evaluated primarily by recall, because any code absent from \(C_v\) is unrecoverable by downstream adjudication.

The original RASC system used SAPBert title embeddings to retrieve the \(K=10\) nearest prior value sets. We replace this title retriever with Qwen3-Embedding-0.6B. Let \(f(t)\in\mathbb{R}^d\) be the Qwen3 embedding of a value set title \(t\) after unit normalization, so \(\|f(t)\|_2=1\). For a source value set \(u\) with title \(t_u\), source sets are ranked by
\[
s(v,u)= f(t_v)^\top f(t_u),
\]
which is cosine similarity because the embeddings are normalized. We retrieve the \(K=15\) highest-scoring source sets,
\[
R_K(v)=\operatorname{TopK}_{u\in\mathcal{V}_{\mathrm{train}}} s(v,u),
\]
and form the initial source-derived pool
\[
C^{\mathrm{src}}_v=\bigcup_{u\in R_K(v)}Y_u.
\]
Duplicate \((\mathrm{code}, \mathrm{system})\) pairs are collapsed. We use \(K=15\) rather than the original \(K=10\) because the downstream objective is high candidate recall and because the stage-2 selector is responsible for rejecting false positives introduced by a broader pool.

\subsection{Vocabulary-Aware Expansion}
Semantic retrieval over prior value sets can miss target codes that are structurally close to retrieved codes but absent from the retrieved source sets themselves. To address this, we define a vocabulary-specific expansion operator. For each expansion-capable vocabulary \(m\), let \(G_m=(U_m,E_m)\) be a graph over codes in that vocabulary. The edge relation encodes clinically meaningful local structure: child and sibling links for ICD-10-CM, relation-neighbor links for RxNorm, and relation-neighbor links for SNOMED-CT. For a seed code \(c\in U_m\), let \(N_m(c)\) denote the allowable neighborhood under this graph.

For a target value set \(v\), expansion is applied to the source-derived pool by
\[
C^{\mathrm{exp}}_v
= C^{\mathrm{src}}_v \cup
\bigcup_{m\in\mathcal{M}_{\mathrm{exp}}}
\bigcup_{c\in C^{\mathrm{src}}_v\cap U_m}
\operatorname{Limit}_m\!\left(N_m(c)\right),
\]
where \(\mathcal{M}_{\mathrm{exp}}=\{\mathrm{ICD\mbox{-}10\mbox{-}CM},\mathrm{RxNorm},\mathrm{SNOMED\mbox{-}CT}\}\). ICD-10-CM expansion uses child and sibling links from the official ICD-10-CM hierarchy artifact. RxNorm expansion uses undirected relation-neighbor edges from official RxNorm files, retaining active English RxNorm atoms and relation attributes such as ingredient, precise ingredient, part-of, tradename, reformulation, and dose-form-group links. SNOMED-CT expansion uses a precomputed same-vocabulary lexical-neighbor graph over codes observed in the local VSAC corpus rather than an asserted complete SNOMED-CT hierarchy.

The operator \(\operatorname{Limit}_m\) denotes deterministic truncation used to keep pools reviewable. For each vocabulary, seed codes are processed in candidate-pool order. For each seed, modes are processed in the configured order: ICD-10-CM uses child then sibling, while RxNorm and SNOMED-CT use neighbor. At most three new codes are added for any one seed; once 60 new codes have been added for a vocabulary for a target value set, expansion for that vocabulary stops. Thus the per-seed cap is applied locally and the per-vocabulary cap is the global stopping rule. Expansion is intentionally restricted to common vocabularies for which local structural expansion is clinically plausible and simple. 

Candidate rows retain provenance features indicating whether a code came from source-set retrieval, vocabulary expansion, direct code-display retrieval, or more than one route. These features are not used to score pool recall, but they are exposed to the stage-2 selectors and retained for audit.

\subsection{Code-Display Rescue Retrieval}
The source-set retrieval step can fail when a target title is phrased differently from titles in the training corpus. This failure mode is especially relevant under publisher shift. To introduce evidence that does not depend on retrieving a similar source value set, we add a code-display retrieval arm over the same expansion-capable vocabularies. For each vocabulary \(m\in\mathcal{M}_{\mathrm{exp}}\), code display strings are indexed with BM25~\citep{robertson2009probabilistic}. Given target title \(t_v\), the top 10 display matches per vocabulary define a seed set \(D_v\). The final candidate pool is then
\[
C_v
= C^{\mathrm{exp}}_v \cup D_v \cup
\bigcup_{m\in\mathcal{M}_{\mathrm{exp}}}
\bigcup_{c\in D_v\cap U_m}
\operatorname{Limit}_m\!\left(N_m(c)\right).
\]
For the final pool used in all stage-2 adjudication experiments, code-display seeds are added for every target value set rather than only when source-set retrieval appears weak. The resulting pool \(C_v\) is the feasible output set for both SAPBert and GPT-5 adjudication.

\subsection{Candidate Adjudication Models}
We evaluate two stage-2 selectors on the same expanded candidate pool. A selector assigns each candidate \(c\in C_v\) either an inclusion score \(g(v,c)\in[0,1]\) or a direct inclusion decision. For score-based models, a threshold \(\tau\) induces the predicted value set
\[
\hat{Y}_v(\tau)=\{c\in C_v:g(v,c)\geq \tau\}.
\]
The first selector is the exported SAPBert cross-encoder from the original RASC workflow. We do not retrain this model on the expanded-pool distribution and do not retune its threshold on the expanded pool. This is a deliberately conservative control: it tests whether simply exposing the original classifier to a larger candidate set is sufficient to improve the workflow. It should not be read as the best possible SAPBert result on the expanded-pool distribution, because the larger pool changes the positive-to-negative ratio and would likely require threshold retuning or retraining for a fair model-development comparison.

The second selector is GPT-5~\citep{openai2026gpt5} used as a constrained adjudicator through an OpenAI-compatible Oracle Cloud Infrastructure Generative AI endpoint. We treat this endpoint as access to the named GPT-5 API model, but endpoint routing, content filters, and model-version changes are external service properties rather than controlled experimental variables. The model is presented with one target value set and its candidate pool and must return a JSON list of candidate IDs judged to belong in the value set. It is not asked to generate clinical codes. Every returned item must correspond to a candidate ID supplied by the retrieval system. Each value set is adjudicated in a separate stateless call with no examples and no conversation history.

We refer to this protocol as blinded because we remove fields that could allow memorization or near-duplicate lookup: target OID, publisher or steward, source value-set OIDs, source value-set titles, true code counts, labels, and true-code lists. The prompt includes the target title, optional VSAC description and status, declared code systems, candidate code-system counts, and candidate rows containing candidate ID, code system, code, display string, and retrieval-derived numeric features. These features expose how a candidate entered the pool, such as source-set support, best retrieval rank, lexical or dense retrieval indicators, hierarchy expansion status, and token overlap with the title. The private scoring manifest retains OID, publisher group, and labels, but those fields are never sent to the model.

Because some expanded candidate pools exceed a single prompt, value sets with more than 1,200 candidates are split into deterministic candidate chunks in the same candidate order used by the serialized pool. The model sees the same target metadata for each chunk and only the candidates in that chunk. Final predictions for a value set are the union of selected candidate IDs across chunks, and a value set is scored only after every chunk for that value set has completed successfully. This chunking strategy preserves the candidate-level inclusion task but prevents the model from comparing candidates across chunks; we return to this limitation below. The full GPT-5 adjudication run completed all 3,744 test value sets, allowing the same all, in-distribution, and held-out-publisher stratification used for the pool-construction analysis.

\subsection{Evaluation}
Evaluation follows the value-set-level convention used in RASC. For a value set \(v\), let \(Y_v\) denote the gold expansion, \(C_v\) the retrieved candidate pool, and \(\hat{Y}_v\) the set selected by a stage-2 model. Candidate-pool recall is the fraction of the gold expansion exposed to the selector:
\[
\mathrm{PoolRecall}(v) = \frac{|Y_v \cap C_v|}{|Y_v|}.
\]
For adjudication, precision, recall, and F1 are computed separately for each value set against the full gold expansion \(Y_v\) and then macro-averaged across value sets. Therefore the reported macro F1 is the mean of per-value-set F1 values, not the harmonic mean of the reported macro precision and macro recall. We also report pool-restricted recall, the fraction of true codes present in the pool that are selected by a stage-2 selector. This separates adjudication failures from retrieval failures:
\[
\mathrm{Recall}_{\mathrm{full}}(v) = \frac{|\hat{Y}_v \cap Y_v|}{|Y_v|}, \qquad
\mathrm{Recall}_{\mathrm{pool}}(v) = \frac{|\hat{Y}_v \cap Y_v|}{|Y_v \cap C_v|}.
\]
Micro-averaged metrics are computed by aggregating true positives, false positives, and false negatives over all candidate decisions. We report ID and OOD strata using the original RASC publisher split. The OOD stratum consists entirely of the held-out Clinical Architecture and CSTE Steward value sets.

\section{Results}
\subsection{Candidate-Pool Recall}
The first experiment measures whether the retrieval stage exposes more of the gold value set to any downstream selector. Table~\ref{tab:pool} reports full-test candidate-pool results as a stage-wise optimization of the RASC retrieval bottleneck. The original SAPBert RASC retriever sets the baseline recall ceiling at 0.553. Replacing the title encoder with Qwen3 and adding vocabulary-aware expansion raises full-test pool recall to 0.654, showing that a stronger semantic retriever plus structurally local expansion recovers many codes that are absent from the original top-\(K\) source-set union. Adding code-display retrieval raises full-test pool recall further to 0.730 and held-out publisher recall from 0.543 to 0.655.

The retrieval-only precision and F1 in Table~\ref{tab:pool} should not be interpreted as final system performance. They correspond to the hypothetical classifier that includes every candidate in the pool. As in RASC, these are macro-averaged value-set metrics; the F1 column is the mean of per-value-set F1 values and is therefore not equal to the harmonic mean of the displayed macro precision and macro recall. The decrease in retrieval-only precision is expected because the stage-1 objective is to reduce false negatives before classification. In the RASC workflow, a false positive introduced by retrieval can still be rejected by the selector, but a true code absent from the pool cannot be recovered downstream.

\begin{table}[t]
\centering
\caption{Full-test candidate-pool performance. Precision and F1 here are retrieval-only macro metrics that predict every candidate as included; recall is the downstream upper bound. Macro F1 is averaged per value set and is not computed from the displayed macro precision and recall.}
\label{tab:pool}
\begin{tabular}{lrrrr}
\toprule
Pool construction & Precision & Recall & F1 & Mean pool size \\
\midrule
Original RASC retrieval~\citep{mukherjee2026retrieveclassifycorpusgroundedautomation} & 0.092 & 0.553 & 0.136 & -- \\
Qwen3 retrieval + vocabulary expansion & 0.0615 & 0.6540 & 0.0997 & 763.7 \\
Qwen3 retrieval + vocabulary expansion + display retrieval & 0.0580 & 0.7298 & 0.0957 & 791.0 \\
\bottomrule
\end{tabular}
\end{table}

\begin{table}[t]
\centering
\caption{Full-test pool recall by publisher regime. ID denotes non-held-out publishers; OOD denotes the held-out Clinical Architecture and CSTE Steward publishers.}
\label{tab:pool-subgroup}
\begin{tabular}{lrrr}
\toprule
Pool construction & All & ID & OOD \\
\midrule
Qwen3 retrieval + vocabulary expansion & 0.6540 & 0.8354 & 0.5428 \\
Qwen3 retrieval + vocabulary expansion + display retrieval & 0.7298 & 0.8518 & 0.6550 \\
\bottomrule
\end{tabular}
\end{table}

\subsection{Stage-2 Adjudication on the Expanded Pool}
The second experiment asks whether the higher-recall pool can be converted into a better selected value set. Table~\ref{tab:llm} reports stage-2 adjudication on the same expanded pool used in Table~\ref{tab:pool-subgroup}. The SAPBert cross-encoder does not automatically benefit from the larger candidate set. On the expanded pool it achieves full-test macro F1 of 0.287, slightly below the original RASC full-test SAPBert cross-encoder result of 0.298, with held-out-publisher macro F1 of 0.233. This is consistent with the expected difficulty of the expanded-pool setting: the pool contains more true positives, but it also contains substantially more plausible false positives.

GPT-5 adjudication over the same expanded pool produces a different stage-2 tradeoff. Full-test macro F1 rises to 0.549, with macro precision 0.625 and macro recall 0.613. The in-distribution and held-out-publisher results are both substantially higher than the SAPBert selector on the same pool. The OOD result is especially important: held-out-publisher macro F1 increases from 0.233 with SAPBert to 0.533 with GPT-5, and held-out-publisher macro precision increases from 0.229 to 0.642 while macro recall increases from 0.415 to 0.564. This does not imply that GPT-5 has solved value-set authoring, but it shows that the retrieval-constrained adjudication formulation can preserve much more of the stage-1 recall ceiling under publisher shift than the original cross-encoder does when applied unchanged.

\begin{table}[t]
\centering
\caption{Stage-2 adjudication on the same expanded pool. Metrics are value-set-level macro averages unless marked micro. The SAPBert cross-encoder is the exported RASC classifier applied to the expanded pool without retraining; GPT-5 is the leakage-controlled constrained adjudicator.}
\label{tab:llm}
\resizebox{\textwidth}{!}{%
\begin{tabular}{llrrrrrr}
\toprule
Selector & Group & \(n\) & Macro P & Macro R & Macro F1 & Pool-restricted R & Micro F1 \\
\midrule
SAPBert CE & All & 3,744 & 0.269 & 0.499 & 0.287 & 0.598 & 0.342 \\
SAPBert CE & ID & 1,423 & 0.333 & 0.636 & 0.374 & 0.676 & 0.423 \\
SAPBert CE & OOD & 2,321 & 0.229 & 0.415 & 0.233 & 0.550 & 0.239 \\
GPT-5 & All & 3,744 & 0.625 & 0.613 & 0.549 & 0.757 & 0.485 \\
GPT-5 & ID & 1,423 & 0.598 & 0.692 & 0.574 & 0.760 & 0.526 \\
GPT-5 & OOD & 2,321 & 0.642 & 0.564 & 0.533 & 0.755 & 0.434 \\
\bottomrule
\end{tabular}
}
\end{table}

Table~\ref{tab:llm} also quantifies the remaining headroom within the retrieved pool. Across the full test set, mean pool recall is 0.730 while full macro recall after GPT-5 adjudication is 0.613. Pool-restricted recall is 0.757, indicating that the selector recovers many of the true codes that retrieval exposes but still misses a meaningful fraction of recoverable positives. The held-out-publisher pattern is similar: mean OOD pool recall is 0.655, GPT-5 full OOD macro recall is 0.564, and OOD pool-restricted recall is 0.755. This supports a practical decomposition. Higher-recall pool construction remains valuable even when the downstream selector is a strong LLM, and adjudication quality must be evaluated separately from the retrieval ceiling.

\subsection{Comparison to Original RASC}
The third experiment places the full workflow in context with the original RASC results. Table~\ref{tab:comparison} compares the main methods on value-set-level macro metrics. Relative to the original RASC SAPBert cross-encoder, the full workflow increases macro F1 from 0.298 to 0.549, an absolute gain of 0.251 and a relative gain of approximately 84\%. Relative to SAPBert applied to the same expanded pool, GPT-5 increases macro F1 from 0.287 to 0.549. This same-pool comparison is the stronger stage-2 evidence, because it holds the retrieval ceiling fixed. The OOD comparison is even sharper: on held-out publishers, GPT-5 achieves macro F1 0.533 on the expanded pool, compared with 0.233 for SAPBert on that pool.

\begin{table}[t]
\centering
\caption{Macro value-set-level comparison. Original RASC results are full-test numbers reported by Mukherjee et al. Expanded-pool SAPBert and GPT-5 are full-test results from the present experiments.}
\label{tab:comparison}
\begin{tabular}{lrrrr}
\toprule
Method & Evaluation set & Precision & Recall & F1 \\
\midrule
RASC retrieval-only & full test & 0.092 & 0.553 & 0.136 \\
RASC SAPBert cross-encoder & full test & 0.272 & 0.483 & 0.298 \\
GPT-4o zero-shot generation & full test & 0.169 & 0.100 & 0.105 \\
Expanded pool + SAPBert cross-encoder & full test & 0.269 & 0.499 & 0.287 \\
Expanded pool + GPT-5 adjudication & full test & 0.625 & 0.613 & 0.549 \\
\bottomrule
\end{tabular}
\end{table}

\section{Discussion}
These results support two claims. First, the RASC benchmark is a useful reusable evaluation harness. It exposes a clinically realistic stress test with a large terminology universe, exact code-level gold labels, held-out publisher shift, and a clean decomposition between retrieval recall and adjudication quality. This makes it well suited for evaluating new embedding models, vocabulary expansion strategies, and constrained LLM selectors.

Second, constrained LLM adjudication appears much more promising than unconstrained clinical code generation. The original RASC GPT-4o baseline attempted to generate value-set codes directly and achieved low recall with frequent invalid codes. In contrast, GPT-5 adjudication is not permitted to invent codes: it selects from a retrieved pool. This preserves the corpus-grounded safety property of RASC while allowing a stronger model to perform semantic inclusion judgments that the SAPBert cross-encoder may not capture.

The held-out-publisher results are central to the evaluation. Stage 1 improves the OOD recall ceiling by adding code-display rescue retrieval to vocabulary-aware expansion, raising held-out-publisher pool recall from 0.543 to 0.655. Stage 2 then determines whether the system can use that noisier pool. SAPBert applied unchanged to the expanded pool selects too many false positives and misses many recoverable positives, yielding OOD macro F1 0.233. GPT-5 adjudication raises OOD macro F1 to 0.533 on the same pool. The remaining gap between OOD mean pool recall 0.655 and OOD macro recall 0.564 is still clinically meaningful, but it is no longer dominated by the severe precision collapse observed with SAPBert on the expanded pool.

The result should also be interpreted as decision support rather than autonomous value-set publication. Even with constrained candidate pools, the system can miss codes absent from retrieval and can accept plausible but stewardship-inappropriate codes. Its most immediate role is to reduce the human review burden: a curator can inspect a ranked or selected candidate set with provenance and retrieval features rather than search the full terminology universe. Cost and latency are non-trivial for API-based GPT-5 adjudication, so a deployed system would likely combine high-recall pooling, cheaper first-pass filtering, and selective LLM review for ambiguous or high-impact concepts.

\section{Limitations}
The main limitation is that the evaluation remains benchmark-based rather than prospective curator evaluation. We do not claim clinical correctness beyond the VSAC expansion labels. VSAC value sets themselves may contain omissions, stewardship-specific modeling choices, or outdated terminology versions. The decision-support claim should therefore be tested in a prospective curator study before deployment claims are made.

Several model-comparison limitations remain. The SAPBert comparison uses the exported original classifier without retraining or threshold retuning on the expanded-pool distribution. This is appropriate as a control for whether the old stage-2 model automatically benefits from a higher-recall pool, but it is not a claim that no retrained cross-encoder, Qwen3 reranker, MedCPT cross-encoder, or threshold-retuned SAPBert model could do better. We also do not report prompt ablations for GPT-5, such as removing retrieval-derived features or adding potentially leaky identifiers, because those would require new experimental runs. The reported comparisons are point estimates without bootstrap confidence intervals or statistical tests; the main F1 differences are large, but uncertainty quantification would be useful for smaller ID/OOD and component-level comparisons.

There are also practical limitations. Large candidate pools are chunked deterministically when they exceed 1,200 candidates, and the final prediction is the union of selected candidates across chunks. This makes the task feasible but prevents the model from comparing all candidates for a value set in a single context. The LLM adjudication protocol uses a proprietary GPT-5 API model accessed through an OpenAI-compatible Oracle Cloud Infrastructure Generative AI endpoint; reproducibility depends on endpoint routing, model-version stability, service availability, and content-filter behavior. Cost and latency are non-trivial because each value set requires one or more API calls over hundreds of candidate rows. We did not record standardized token-level cost and wall-time measurements in the manuscript artifacts, so deployment cost should be measured separately before any production pilot.

\section{Conclusion}
Retrieval-constrained value set authoring can be improved by optimizing candidate-pool construction and candidate adjudication separately. Stage 1 raises the full-test recall ceiling from 0.553 in the original RASC retrieval baseline to 0.730; on held-out publishers, the expanded pool reaches 0.655 recall. Stage 2 then converts the larger pool into better end-to-end predictions: GPT-5 adjudication reaches full-test macro F1 0.549 and held-out-publisher macro F1 0.533, compared with 0.287 and 0.233 for SAPBert on the same expanded pool. These results preserve the key safety constraint that outputs must come from a retrieved candidate pool while substantially improving both in-distribution and OOD performance. The evidence suggests that retrieval-constrained LLM adjudication is a strong direction for clinical value set authoring and that the RASC benchmark provides a useful testbed for this class of systems.

\bibliographystyle{unsrtnat}
\bibliography{rasc_plus_ml4h}

\appendix
\section{LLM Adjudication Prompt}
The GPT-5 adjudication call used a fixed system instruction and a JSON user payload containing one blinded value-set packet. The system instruction was: ``You are adjudicating clinical value-set candidate codes. Task: Given one target value-set description and a candidate code pool, select every candidate that should belong in the value set. Use only the target metadata and candidate code information provided in this prompt. Do not use hidden knowledge of VSAC OIDs, publishers, stewards, or source value sets. Do not explain your reasoning. Return strict JSON only, with this schema: \texttt{\{"selected\_candidate\_ids":["c000001","c000017"],"notes":""\}}. If no candidates should be included, return \texttt{\{"selected\_candidate\_ids":[],"notes":""\}}. Candidate IDs must be copied exactly from the prompt.''

The user payload was a compact JSON object with three top-level fields. The \texttt{packet\_id} field contained an anonymized experiment identifier with no VSAC OID. The \texttt{target} field contained the value-set title, optional description and status, declared code systems, candidate code-system counts, and candidate pool size. The \texttt{candidates} field contained one object per candidate with candidate ID, code system, code, display string, and retrieval-derived features. The prompt did not include labels, true-code lists, true-code counts, target OID, publisher, steward, source value-set OIDs, or source value-set titles.

\end{document}